%% file: main.tex
\documentclass[letterpaper, 10 pt, conference]{ieeeconf}  

\IEEEoverridecommandlockouts                              

\usepackage{graphics} 
\usepackage{epsfig} 
\usepackage{mathptmx} 
\usepackage{times} 
\usepackage{amsmath} 
\usepackage{amssymb}  
\usepackage{float}
\usepackage{afterpage}
\usepackage{changepage}
\usepackage{multicol}

\usepackage[font=small,labelfont=bf]{caption}

\usepackage{subcaption}
\usepackage{adjustbox}
\usepackage[usenames,dvipsnames]{xcolor}
\usepackage{colortbl}
\usepackage{lipsum} 
\usepackage{microtype}
\usepackage{booktabs}
\usepackage{graphicx}
\usepackage{multirow}
\usepackage{pifont}
\usepackage{url}
\usepackage{xparse}
\usepackage{xspace}
\usepackage{nicefrac}
\usepackage{enumerate}
\usepackage{pifont}%
\usepackage{setspace}
\usepackage{wrapfig}
\usepackage{xfrac}
\usepackage{tikz}

\usepackage{stfloats}
\usepackage[most]{tcolorbox}
\usepackage{hyperref}

\definecolor{MyDarkRed}{rgb}{0.8,0.02,0.02}
\definecolor{MyDarkGreen}{rgb}{0.02,0.6,0.02}
\definecolor{MyPurple}{rgb}{0.6,0.1,.9}

\hypersetup{
 colorlinks=True,
 linkcolor=blue,
 citecolor=blue,
 citebordercolor=blue,
 urlcolor=blue}

\newcommand{\method}{\texttt{LCB}}
\newcommand{\act}{{\texttt{<ACT>}}}

\title{\LARGE \bf From LLMs to Actions: Latent Codes as Bridges in \\ Hierarchical Robot Control}

\author{
Yide Shentu$^{*}$ \qquad 
Philipp Wu$^{*}$ \qquad 
Aravind Rajeswaran \qquad 
Pieter Abbeel \\
[.5ex]
*Equal contribution \\
University of California, Berkeley \qquad 
}

\begin{document}

\makeatletter
\let\@oldmaketitle\@maketitle%
\renewcommand{\@maketitle}{\@oldmaketitle%
    \centering
    \vspace*{-0.8ex}
    \includegraphics[width=0.99\linewidth]{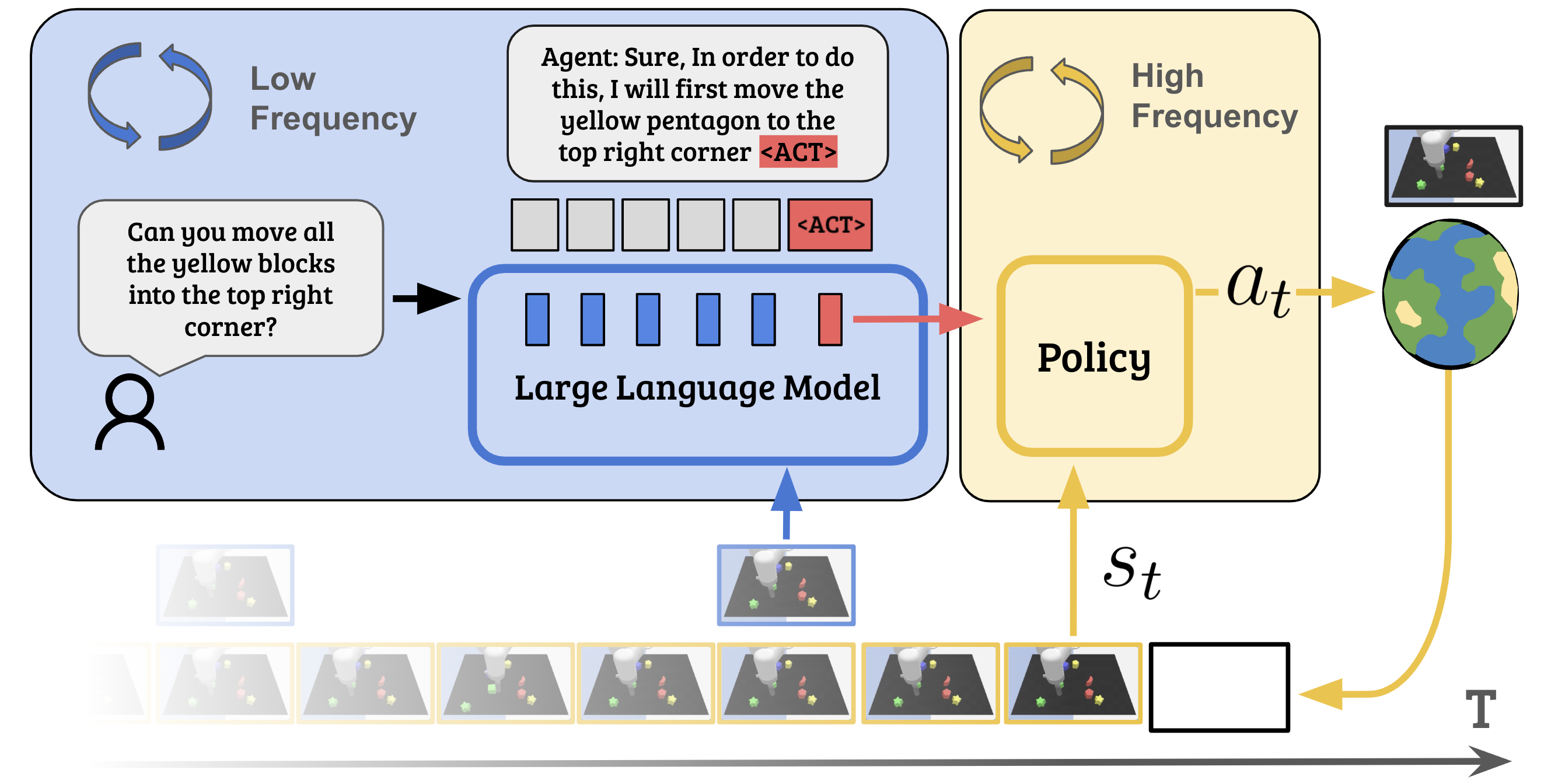}
    \vspace*{-0.5ex}
    \captionof{figure}{\textbf{Illustration of our proposed Latent Code as Bridges architecture.}
    Given a high-level task description and the observation, a Large Language Model (LLM) generates a textual description of an action and an \act~token. The feature embedding from the \act~token's last layer serves as a high-level latent goal for the downstream policy network. Our modular hierarchical approach synergies the LLM's high-level reasoning with the pre-trained policy's responsive low-level control, addressing the limitations of direct action output by monolithic LLMs. Unlike methods that  using a large LLM to directly output agent actions \cite{brohan2023rt2}, our approach can run the LLM reasoning and action policy execution loops asynchronously, mirroring human task execution with immediate low-level feedback when interacting with the physical world and slower, deliberate reasoning when considering longer term planning. At test time, the action policy frequently updates actions based on environment changes, while the LLM updates are less frequent, enabling efficient, real-world inference.}
    \label{fig:teaser}
}
\makeatother

\maketitle
\thispagestyle{empty}
\pagestyle{empty}

\input{sections/0-abstract}
\setcounter{figure}{1}
\input{paper.tex}




\section*{ACKNOWLEDGMENTS}
Yide Shentu is supported in part by InnoHK Centre for Logistics Robotics and ONR MURI N00014-22-1-2773. Philipp Wu is supported in part by the NSF Graduate Research Fellowship Program and Multidisciplinary University Research Initiative (MURI) award by the Army Research Office (ARO) grant No. W911NF-23-1-0277. We thank Xinyang Geng and Fangchen Liu for valuable discussions regarding \method.

\bibliography{references}
\bibliographystyle{IEEEtran}

\input{appendix}

\end{document}

%% file: sections/0-abstract.tex
\begin{abstract}
Hierarchical control for robotics has long been plagued by the need to have a well defined interface layer to communicate between high-level task planners and low-level policies. With the advent of LLMs, language has been emerging as a prospective interface layer. However, this has several limitations. Not all tasks can be decomposed into steps that are easily expressible in natural language (e.g. performing a dance routine). Further, it makes end-to-end finetuning on embodied data challenging due to domain shift and catastrophic forgetting. We introduce our method -- Learnable Latent Codes as Bridges (LCB) -- as an alternate architecture to overcome these limitations. \method~uses a learnable latent code to act as a bridge between LLMs and low-level policies. This enables LLMs to flexibly communicate goals in the task plan without being entirely constrained by language limitations. Additionally, it enables end-to-end finetuning without destroying the embedding space of word tokens learned during pre-training. Through experiments on Language Table and Calvin, two common language based benchmarks for embodied agents, we find that \method~outperforms baselines (including those w/ GPT-4V) that leverage pure language as the interface layer on tasks that require reasoning and multi-step behaviors. 
\end{abstract}

%% file: paper.tex
\input{sections/1-introduction}
\input{sections/2-related_works}

\section{Method}
\label{sec:method}

\begin{figure*}[tb]
  \vspace*{2ex}
  \centering
  \includegraphics[width=0.99 \textwidth]{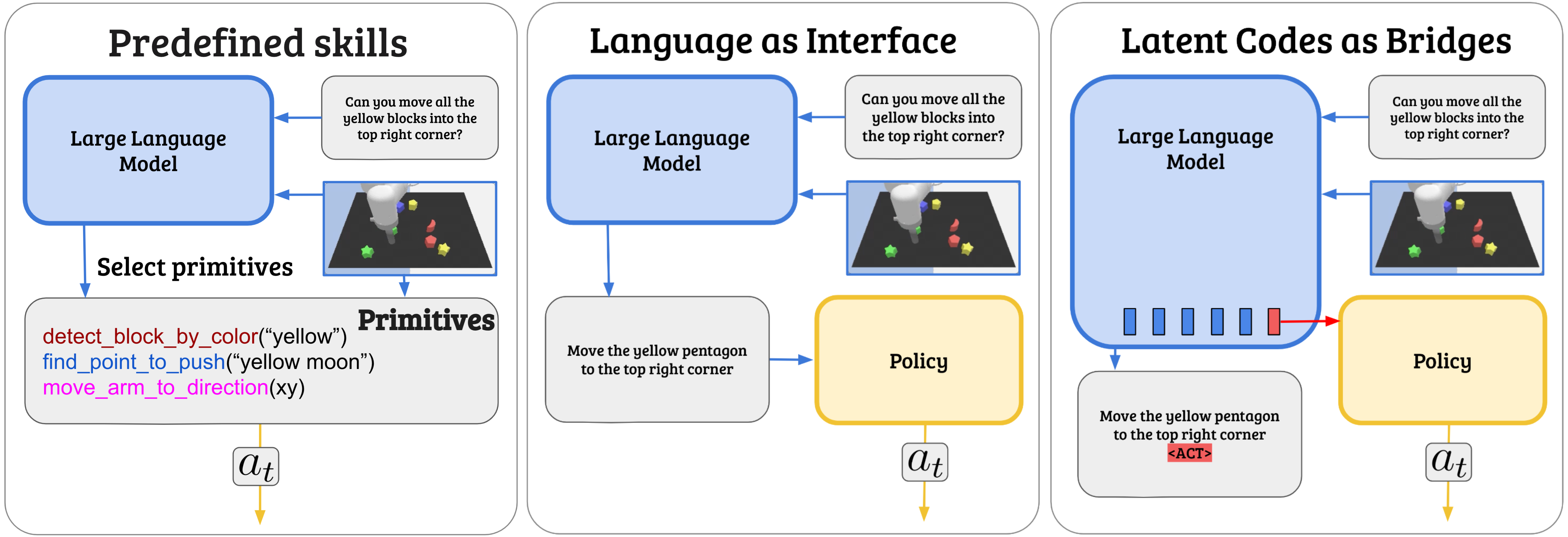}
  \caption{
      A high level architectural comparison of LLM-based hierarchical policies.
      Predefined skills (left) uses a LLM to call predefined primitives. Language as an interface (middle) uses a LLM to output a simple language command, which is then passed into a language conditioned policy. \textbf{\method}~(right) utilizes a latent code as a \textbf{bridge} between the LLM and the low level policy, facilitating hierarchical control and end-to-end learning.
  }
  \label{fig:methods}
\end{figure*}

We wish to develop a hierarchical policy architecture that can enable robots to perform a variety of manipulation tasks when provided with free-form language descriptions. Specifically, we seek an architecture that can handle low-level actions for fine-grained or contact-rich tasks (e.g. pushing, 6D object manipulation) while also having the capability to reason and plan without any external step-by-step instructions. Before we present our architecture for this purpose, we first survey two other families of approaches and their deficiencies, which provides the intuition and basis for our method. These approaches are shown in \autoref{fig:methods}.\\

\noindent \textbf{LLMs Leveraging Predefined Skills} First we can consider a hierarchical approach where LLMs perform high-level task planning by calling a set of pre-defined skills or APIs~\cite{Ahn2022-gz, Liang2022-vh}. These APIs (e.g. \texttt{go-to}, \texttt{push}) are described and provided to the LLM as part of the main prompt. This approach suffers from two primary drawbacks. Firstly, for an LLM to plan with skills, they need to have \textit{semantics} attached to them that make linguistic sense. Secondly, this constrains the set of skills to a closed vocabulary, and prevents any form of generalization to new skills or capabilities. Furthermore, code-writing proficiency demands a high-quality LLM, a criterion met chiefly by proprietary commercial models such as GPT-4~\cite{liu2024moka}. Additionally, end-to-end fine-tuning is challenging since the LLM cannot adapt or compensate for limited prowess of the low-level skills~\cite{Ahn2022-gz}. \\

\noindent \textbf{Language as Interface} The second class of approaches can leverage \textit{language-conditioned low-level policies} as opposed to a finite set of low-level skills. Such policies can take a simple language command as input (e.g. \texttt{pickup the red block}) and produce actions that can (hopefully) accomplish the task. Since these policies can accept free-form text as input, at least theoretically, they have the capability to generalize to new instructions. Furthermore, they are amenable to end-to-end fine-tuning from high-level instructions, through an LLM, to the language conditioned policy, and ultimately the action. Nevertheless, this class of approaches also suffer from key limitations. Firstly, not all high level tasks can be decomposed into sub-tasks in simple language. For example, imagine trying to describe step-by-step instructions to make a robot dance to a song. Secondly, end-to-end fine-tuning with such an architecture can destroy planning and reasoning capabilities that the LLM originally had~\cite{luo2023empirical}. \\

\noindent \textbf{Latent Codes as a Bridge (Ours)} Finally we describe our method which can overcome the key limitations outlined above. Our key insight is that we can introduce an additional latent code to act as a bridge between the high-level LLM and low-level language conditioned policy. We augment the LLM's tokenizer by adding a specialized \act~token, prompting the model to predict this token in response to actionable questions. The last layer embedding of the \act~token is then utilized as a latent goal for the downstream policy network. This learnable \act~token's embedding facilitates the transmission of abstract goals and nuances to the low-level policy -- details that are not easily conveyed through language alone. Furthermore, by using this additional learnable token, we preserve the embedding space for language tokens, thus preventing any catastrophic forgetting during end-to-end fine-tuning. We describe more specific details of our architecture and implementation below. 

\subsection{Architecture and Implementation Details of \method}

\method~unifies the capabilities of a slow but powerful pretrained Multimodal Large Language Models (LLMs) with a fast and simpler decision-making policies to create a model that ingests vision and language inputs to output low-level actions. This integration involves a two-component system: a pretrained LLM, denoted as $f_{\phi}$, and a pretrained policy, $\pi_{\theta}$, parameterized by $\phi$ and $\theta$ respectively. The LLM consists of a text only large language model and a vision encoder, which projects images into the text only large language models embedding space, facilitating a multimodal understanding of textual and visual inputs. In this work, we leverage LLaVA\cite{liu2023visual} as our pretrained LLM.
$f_{\phi}$ takes in text tokens $x_{txt}$ and images $x_{img}$ and outputs text tokens. The pretrained policy $\pi_{\theta}$ takes as input environment observations at the current time step $o_{t}$, with conditioning latent $z$, and outputs the action at the current time step $a_t$. 

We introduce an additional \act~token into the vocabulary of the language model, which is a special token that enables the language model to generate an action embedding to control the lower level policy. The model is trained to output \act~tokens when executable requests are provided to the model. We extract out the last-layer embedding features from the model of at the \act~token, following the approach used in Language Instructed Segmentation Assistant (LISA) \cite{lai2023lisa}. This embedding is projected into the policy latent conditioning space by a linear layer to extract the latent feature $z_{\act}$~which is then fed into the policy $\pi_{\theta}$.

\subsection{Data Processing}

The \method~framework necessitates diverse and strategically curated datasets to make the policy effective for language-guided action execution in varied contexts. We cater the data collection and preprocessing steps towards this goal, creating a small instruction tuning dataset.

We convert in domain text conditioned policy data into the chat format of LLM assistants. Typical language conditioned trajectory datasets contain one language instruction and a list of (observation, action) pairs $[x_{txt}, (o_0, a_0, ..., o_t, a_t, ...)]$ per trajectory. We programmatically generate text data in the format of chat interactions using templates. A simple example of this user-assistant interaction, is ``User: can you help me $x_{txt}$? Assistant: yes, \act." Specific templates for chat data generation are provided in Appendix \ref{append:dataset_details_language_table}. This trains the model to recognize and respond to direct action requests, fostering a conversational interface that seamlessly transitions from dialogue to action.

Moreover, we enrich our training material with additional datasets designed to prompt specific behaviors from the language model. One such data source is reasoning data, where the model is tasked with a more abstract goal and must reason about the scene to accomplish the goal. Such examples are framed within a chat-like interaction, encouraging the model to articulate its reasoning process before executing the \act~command. For example, ``User: $x_{img}$ Can you ${x_{txt}}$? Assistant: I will $x_{goal}$ \act". Where ${x_{txt}}$ does not explicitly specify the target object and location. If ${x_{txt}}$ is ``move the block closest to the bottom right to the block of a similar color", the assistant's response, $x_{goal}$, provides an explanation of the task, such as ``I will move the blue cube on the bottom right to the blue moon".

We also study long-horizon tasks and incorporate training sequences that require the model to plan and execute multiple steps to achieve a goal. This is achieved by defining task stages (start, regular, transition, stop) and incorporating the previous action as context in the language model's input. This strategy trains the model to recognize task progression and adapt its actions accordingly, enabling it to manage tasks with evolving objectives. 
Through this dataset strategy, our model is finely tuned as a versatile tool capable of understanding and executing a wide range of language-guided actions. 

\input{figures/env/figure}

\subsection{Training}

The training of \method~employs a combination of techniques to integrate the LLM and policy components.
We leverage Low Rank Adaptation\cite{hu2022lora} (LoRA)  for fine-tuning the LLM, allowing for more efficient training.
We adopt a cold start approach to policy training, reminiscent of staged training strategies seen in prior works, by first freezing the action decoder and only fine-tuning the language model.
This preliminary phase focuses on aligning the embeddings produced by the LLM with the feature space of the policy.
We find that adding an additional CLIP loss to regularize the latent embedding $z_{\act}$ is necessary, ensuring that the embeddings from the language model remain well aligned with the lower level ground truth text description $g_{txt}$ of the objective for the pre-trained policy.
In total, our loss function is comprised of 3 terms, and can be expressed as follows:

\begin{align}
\mathcal{L} =&\lambda_1 \mathcal{L}_{\text{policy}}(\pi_{\theta}, o_t, a_t, z_{\act}) \\
+&\lambda_2 \mathcal{L}_{\text{LM}}(f_{\phi}, x_{\text{txt}}, x_{\text{img}}) \\
+&\lambda_3 \mathcal{L}_{\text{CLIP}}(z_{\act}, g_{\text{txt}})
\end{align}

\input{figures/bar_plots/figure}

\section{Results}
\label{sec:results}
We systematically evaluated \method~across a diverse set of environments and tasks to demonstrate the efficacy of integrating a pretrained Large Language Model (LLM) with a domain-specific, pretrained low-level policy. Our primary objective was to study the capabilities of the policy, specifically its high-level language understanding and low-level control.
Through our experiments, we aim to answer the following questions:

\begin{itemize}
    \item Does \method~enable learning a bridge between the LLM and the policy more effectively than pure language?
    \item Can \method~leverage the pretrained capabilities of LLMs to solve long horizon tasks by decomposing the high level goals into the step by step latent commands?
    \item Can \method~outperform other baseline methods that leverage close-sourced state of the art LLMs such as GPT-4V?
\end{itemize}

To answer these questions, we study how \method \space performs under various reasoning and long horizon settings in both the Language Table and CALVIN benchmarks. See \autoref{fig:envs} for a visualization of the environments and example tasks.

\subsection{Evaluation on Language Table}
Language Table offers a simulated tabletop environment for executing language-conditioned manipulation tasks \cite{Lynch2022-na}. The environment features a flat surface populated with blocks of various colors and shapes, alongside a robot with 2D action space.  Language Table provides observations in the form of the robot end effector position and third-person camera images. Despite its simplicity, it provides a reproducible and comprehensive environment to study challenges at the interface of high level language and low level contact-rich dynamics and feedback control.

\input{tables/table_og_langtab}

We investigate the benefit of using \method~on the original Language Table benchmark. Here we apply our method using the same dataset that the original language table model was trained on, translating the original language instructions into chat interactions with action tokens as specified in \autoref{sec:method}. As shown in \autoref{tab:langt_success_rates}, with the end to end optimization with the pretrained LLM, the success rate across the benchmark matches or exceeds the baseline Language Table approach. This signifies that \method~is able to seamlessly adapt a pretrained LLM and policy together. We suspect that this is due to the flexibility in the latent representation $z_{\act}$, allowed for by our approach as well as additional capacity afforded my the language model.

We next investigate more complex language tasks that require reasoning and planning capabilities. We collect a small dataset for each capability, training models to compare the following approaches:
\begin{itemize}
    \item \textbf{LangTable:} The original Language Table Policy, as provided by \cite{Lynch2022-na}.
    \item \textbf{LangTable + LLaVA (Frozen):} The combination of the original policy and a non-fine-tuned LLaVA model interfacing through language. We prompt LLaVA to output language commands in the format and style as expected by LangTable.
    \item \textbf{LangTable + GPT-4V:} The integration of LangTable with the state-of-the-art proprietary Vision Language Model (GPT-4V). In order to bootstrap the spatial understanding of GPT-4V, we also incorporate the Set of Marker (SOM) \cite{yang2023setofmark} technique to enhance the GPT-4V's capability. We further include multi-modal few show contexts including language explanation of the tasks and image examples. More details are provided in Appendix \ref{append:gpt_prompt}.
    \item \textbf{LangTable + LLaVA (Fine-tuned):} The original policy augmented by a LLaVA model that has been fine-tuned on the exact language needed for the action policy for the given task.
    \item \textbf{\method{}:} We take a pretrained LLaVA model and the pre-trained LangTable policy and apply \method, learning a latent interface between the two on the respective instruction dataset. 
\end{itemize}

Results for long horizon performance are provided in \autoref{fig:multi_step}. In this task, the agent must sort blocks based on shape or color into a specified corner of the board, requiring a long sequence of actions from which the agent could greatly benefit through high-level planning. We see that \method~ exhibits a competency for handling such tasks, as indicated by the heightened success rates, improving on pure language interface baselines. This is attributable to the method's ability to generate a coherent sequence of latent action embeddings that guide the policy through the task's duration, facilitating a more consistent and accurate alignment with the sequential nature of the task. During evaluation we run the higher level language model at a slower rate than the lower level policy, only updating the language models output every 40 environment steps. We find that this increases computational efficiency without compromising task performance suggesting the effectiveness of the model hierarchy.

Results for reasoning performance are provided in \autoref{fig:reasoning}. Tasks here are of the form ``There is a block that is closest to \{corner\}. Push that block to the other block of the same \{shape/color\}". In order to successfully accomplish this task, the agent must identify which block is located closest to a given corner, identify the relevant property (i.e. shape or color) and consolidate that understanding into an executable instruction. We see that our approach is able to outperform baselines that involve zero-shot prompting as well as naively fine-tuning the language model to output the translated robot task. We see that fine-tuning the language model to output the ground truth language primitive is effective in reaching parity with the oracle language baseline, but that \method~is able to match and even exceed that.

We provide a qualitative assessment of the language output from the various top performing approaches in \autoref{fig:lang_compare}. LangTable + GPT-4V requires heavy prompt engineering and additional string parsing to extract out the final policy. LangTable + LLaVA is effectively fine-tuned by outputting the direct low level text command to the policy, but no longer is able to maintain a chat like interface to the user. In contrast, \method~is able to output an effective embedding for the low level policy while also verbalizing its reasoning. This decouples the low level policy conditioning from the language models text outputs, offering increased flexibility during instruction fine-tuning.

\input{figures/response_comparison/figure}

\subsection{Evaluation on CALVIN}
CALVIN\cite{mees2022calvin} is an open-source simulated benchmark designed for learning long-horizon tasks conditioned by language. The environment features a 7-DOF Franka Emika Panda robotic arm equipped with a parallel gripper, situated at a desk with a variety of articulated furniture and objects for interaction. In each experiment, the robot needs to solve a sequence of complex full 6D manipulation tasks governed by real-world physics and guided by a series of language instructions. Each subtask is paired by a specific language instruction; upon successful completion, the robot proceeds to the next subtask accompanied by a new instruction. CALVIN encompasses four distinct environments A, B, C and D, with a shared set of language instructions and subtasks.

\input{tables/table_calvin_enriched}

In order to demonstrate the generalization capabilities of \method~ cross various environments as well as its ability to comprehend and act upon the same instructions phrased differently in the CALVIN long horizon full 6D manipulation setting, we compare the following approaches:\\
\begin{itemize}
    \item \textbf{RoboFlamingo (RF):} RoboFlamingo\cite{li2024visionlanguage} adapts OpenFlamingo\cite{awadalla2023openflamingo} by fine-tuning solely the cross-attention layer to directly output actions, thus maintaining its language comprehension. However, this approach requires executing the entire LLM anew with each progression to a subsequent state, leading to inefficiencies.
    \item \textbf{3D Diffusion Actor (3DDA):} Incorporating a diffusion policy with 3D scene representation and CLIP\cite{radford2021learning} language embedding, the 3D Diffusion Actor \cite{ke20243d} sets the current SOTA on the Calvin benchmark when provided with standard language instruction inputs. However, a notable limitation stems from the constraints of the CLIP text model it employs. 3DDA can not generalize well on language instruction outside of its training distribution. 
    \item \textbf{\method:} \method \space for Calvin integrates a pre-trained LLaVA\cite{liu2023visual} as the Multimodal Large Language Model backbone with a pre-trained 3D Diffusion Actor serving as the action policy. This combination leverages the SOTA capabilities of the 3D Diffusion Actor to achieve a synergistic effect: \method\space for Calvin excels in both language comprehension and low-level manipulation. Since RoboFlamingo runs the entire LLM on every environment step, in order to make a fair comparison, we also run the LLM part of \method~ synchronously with the downstream policy, although we notice no significant performance difference for Calvin.  
\end{itemize}

~\autoref{table:resultcalvin} presents results for the CALVIN long-horizon, language-conditioned benchmark. In this setting, the robot executes a series of tasks in unfamiliar environments based on novel GPT-4 enriched\cite{awadalla2023openflamingo} instructions not encountered during training. The experimental outcomes demonstrate our approach's distinct advantage over baseline methods. \method \space significantly surpasses all baselines in terms of task success rate at every stage and in average completed trajectory length.

\section{Conclusion}
\label{sec:conclusion}
In this work, we introduce a novel approach, Latent Codes as Bridges, or \method, that combines the abstract reasoning capabilities of large language models with low-level action policies. Our methodology does not merely stack these capabilities as in prior works but integrates them in an end-to-end fashion through a learned latent interface. The empirical evidence from our evaluations on the Language Table and CALVIN benchmarks shows the model's adeptness in interpreting and executing various reasoning and long horizon objectives. The flexibility and effectiveness of the hierarchy enabled by \method~shows promise for real world robotic applications.

%% file: sections/1-introduction.tex
\section{INTRODUCTION}
\label{sec:intro}

The field of robotics has long oscillated between two predominant architectural paradigms for enabling agents to solve complex tasks. At one end of the spectrum, we have seen \textit{\bf modular hierarchical policies}~\cite{liu2024moka} for control that leverage rigid layers like symbolic planning, trajectory generation, and tracking. On the other end are \textit{\bf end-to-end policies}~\cite{Levine2015EndtoEndTO} that directly map sensory observations to actions through high-capacity neural networks. This dynamic history reflects the ongoing quest to reconcile the logical human-like reasoning with the flexible dexterity of human motor control.

The advent of \textit{large language models} (LLMs)~\cite{openai2023gpt4, touvron2023llama} and their remarkable language interpretation and reasoning capabilities have reignited interest in hierarchical control architectures. Recent works~\cite{Ahn2022-gz,Huang2022-gl,Liang2022-vh} have leveraged LLMs and \textit{Multimodal Large Language Model} (abbreviated as LLM in this paper unless specified otherwise) in place of high-level symbolic planners, enabling impressive results like mobile rearrangement of objects based on open-vocabulary instructions. Despite these advances, the core deficiencies of hierarchical architectures remain -- namely the need for a set of clearly defined control primitives and an interface between layers in the hierarchy. For example, LLMs leverage the semantic meaning of action verbs to coordinate low-level primitives like \textit{go-to}, \textit{pick}, \textit{place} etc. However, we humans perform a variety of movements with our body that contribute to our dexterity and daily function, yet \textbf{cannot be easily described using language.}

In this backdrop, we present \textbf{L}atent \textbf{C}odes as \textbf{B}ridges, or \method, a new policy architecture for control that combines the benefits of modular hierarchical architectures with end-to-end learning (see Fig.~\ref{fig:teaser} for an illustration). Specifically, not only can \method~directly leverage LLMs for high-level reasoning and pre-trained skills/policies for low-level control, but it can also improve these components with end-to-end learning to transcend their initial capabilities. 
This is achieved by learning an \act~token at the interface layer which can modulate the low-level policies. As a result of this choice, \method~can overcome the inherent limitations of solely relying on language as the interface layer, since several behaviors are hard to describe in language. Secondly, by leveraging a separate \act~token, we do not destroy the core language generation and reasoning capabilities of the LLM during finetuning. 
We test \method~on a series of long-horizon and reasoning tasks in Language Table~\cite{Lynch2022-na} and Calvin~\cite{mees2022calvin}, two common language based benchmarks for embodied agents. We find that \method ~considerably outperforms baselines that leverage LLMs to sequence low-level skills using pure language as the interface layer. See our \href{https://fredshentu.github.io/LCB_site}{website} for more. 

%% file: sections/2-related_works.tex
\section{RELATED WORK}
\label{sec:related}

\noindent \textbf{Hierarchical Control with LLMs}
The proliferation of LLM technology, coupled with their capability to interpret user prompts and perform reasoning, has led to growing interest in utilizing LLMs for robotics~\cite{Zeng2023LargeLM, Vemprala2023ChatGPTFR}. Of particular notice and relevance are the use of LLMs for high-level reasoning in hierarchical control architectures. Prior work has demonstrated this by leveraging the few-shot prompt capabilities of LLMs~\cite{Huang2022-gl, Ahn2022-gz}, their ability to code and compose functions~\cite{Liang2022-vh, singh2023progprompt}, or their ability to interact with human users through language~\cite{li2023itp}. In contrast to these works that attempt to use LLMs ``as-is'' and compose low-level skills, our work performs end-to-end fine-tuning through learnable latent codes. This includes finetuning some layers of the LLM through LoRA\cite{hu2022lora}. Empirically we show that such finetuning can outperform methods that use LLMs out-of-the-box.

\noindent \textbf{Language Conditioned Imitation Learning}
To leverage LLMs for task planning and reasoning, such models need to be able to call lower-level skills to affect change in the environment. This can be achieve in two ways: (a) by leveraging \textit{semantics} of the skills through language descriptions (e.g. \texttt{go-to}, \texttt{reach} etc.) as described above; or alternatively (b) through language conditioned policies which accept a text description as input to directly produce an action~\cite{Lynch2022-na, Jang2022BCZZT, brohan2023rt2, li2023vision, lynch2021language}. Such policies can typically perform only short horizon tasks and lack the reasoning and planning capabilities often found in LLMs. Our goal in this work is to leverage such ``simple'' or ``primitive'' language-conditioned policies along with LLMs to enable a hierarchical system to perform complex tasks that require multi-step planning and reasoning. 

\noindent \textbf{Large Pre-Trained Models for Embodied Agents} Recent years have witnessed growing interest in robotics to re-use large models originally trained for vision or language applications~\cite{Kirillov2023SegmentA, Vemprala2023ChatGPTFR} or their architectures~\cite{Chen2021DecisionTR, janner2021sequence, wu2023mtm, liu2023masked}. We are also starting to see large models and representations custom trained for robotics~\cite{brohan2023rt2, Nair2022R3MAU, vc2023, yang2024learning}. In our work, we leverage the recent class of Multimodal Large language models~\cite{liu2023visual, zhu2023minigpt, Li2023BLIP2BL} that extend the capability of text only LLMs to interpret other modalities like vision through alignment layers. Specifically, our instantiation of \method~model builds on top of LLaVA~\cite{liu2023visual} and finetunes the model on a simulated dataset of embodied reasoning and long-horizon tasks. As the availability of embodied datasets paired with language annotations grow, we hope that our method can be extended to release generalist models that can be deployed zero shot in new domains. 

%% file: figures/env/figure.tex
\begin{figure}[b!]
  \centering
  \includegraphics[width=1.0\linewidth]{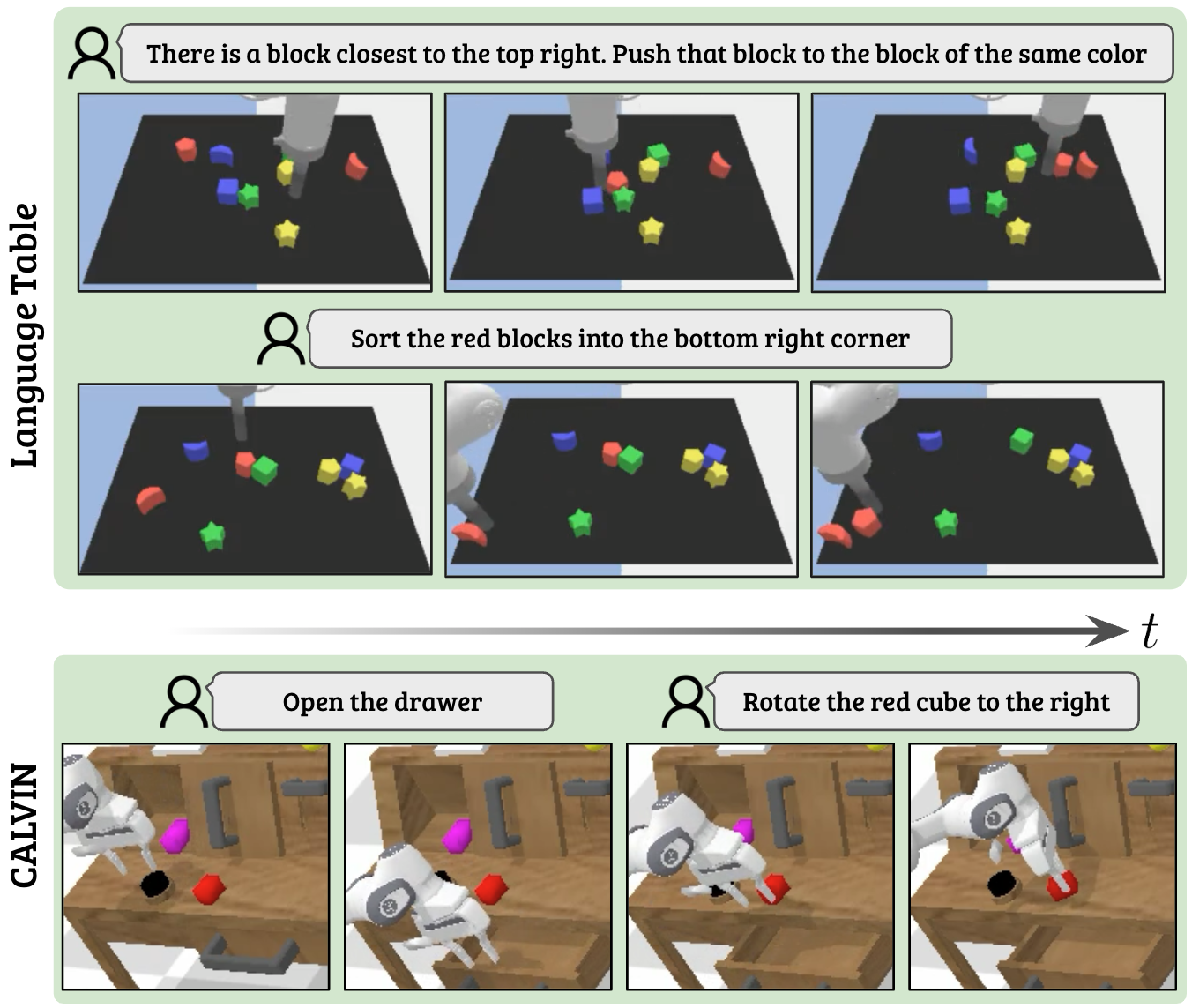}
  \caption{
      A visualization of the two environments along with exemplar tasks that we train and evaluate on. The top depicts the Language Table environment \cite{Lynch2022-na}. We study reasoning tasks (first trajectory) and long horizon tasks (second trajectory). The bottom depicts the CALVIN long horizon benchmark \cite{mees2022calvin}, in which the agent must sequentially accomplish tasks.
  }
  \label{fig:envs}
\end{figure}

%% file: figures/bar_plots/figure.tex
\begin{figure*}[tb]
\vspace*{2ex}
\begin{subfigure}{\textwidth}
    \includegraphics[width=\textwidth]{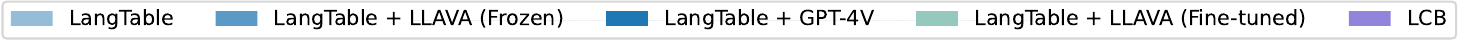}
\end{subfigure}
\hfill
\begin{subfigure}{0.49\textwidth}
    \includegraphics[width=\textwidth]{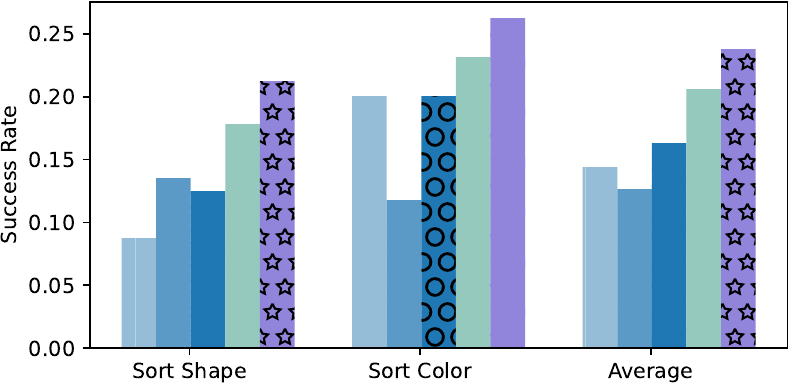}
    \caption{\textbf{Long Horizon} Success rate for the multi-step tasks on Language Table. The task requires shorting some blocks based on color or shape in a given direction. The environment only provides the high level objective to each method. This task requires the policy to have more long term planning capabilities, whether explicitly or implicitly. }
    \label{fig:multi_step}
\end{subfigure}
\hfill
\begin{subfigure}{0.49\textwidth}
    \includegraphics[width=\textwidth]{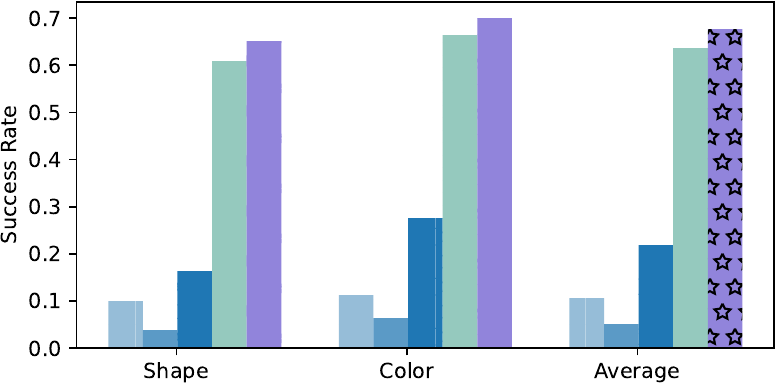}
    \caption{
        \textbf{Reasoning:} Success rate for the reasoning tasks on Language Table. The reasoning task is specified as a variant of "There is a block that is closest to {i.e., top right corner}. Push that block to the other block of the same {shape/color}." This task requires the agent to understand object semantics and spacial relationships.
    }
    \label{fig:reasoning}
\end{subfigure}
\hfill
\caption{
Task success rates on Language table. The tasks are drawn from the higher level Language Table tasks from PALM-E \cite{Driess2023-lm}. LangTable refers to the original language table policy \cite{Lynch2022-na}.  +LLaVA (frozen) refers to composing the original language table with a frozen LLaVA model and few shot prompting. +GPT-4V similarly refers to composing the original policy with GPT-4V. +LLaVA (finetuned) refers to finetuning the LLaVA policy on our mixture dataset on the language only, then composing it with the policy. Our results show that leveraging \method{} is effective on tasks that require additional reasoning and planning. Note that the same model is evaluated between the long horizon and reasoning tasks.
}
\label{fig:barplot}
\vspace*{-1ex}
\end{figure*}

%% file: tables/table_og_langtab.tex
\begin{table}[t]
  \small
  \caption{Comparison on the original Language Table benchmark tasks. LangTable is the original language table policy \cite{Lynch2022-na}. \method~is our method applied only to the original Language Table dataset. We see that \method~can help improve task performance by leveraging the vision language model for feature extraction. The tasks are: Block to Block (B2B), Block to Block Relative Location (B2RL), Seprate (S), Block to Relative Location (B2RL), and Block to Absolute Location (B2AL).}
  \label{tab:langt_success_rates}
  \centering
  \begin{tabular}{l l l l l l c}
    \toprule
    Model &  B2B & B2BRL & S & B2RL & B2AL & \textit{Avg} \\
    \midrule
LangTable & 0.88 & 0.70 & 0.94 & 0.68 & 0.65 & 0.77 \\
\method & 0.90 & 0.66 & 0.99 & 0.73 & 0.71 & \textbf{0.80} \\
    \bottomrule
  \end{tabular}
\end{table}

%% file: figures/response_comparison/figure.tex
\begin{figure}[t!]
  \vspace*{2ex}
  \centering
  \includegraphics[width=1.0\linewidth]{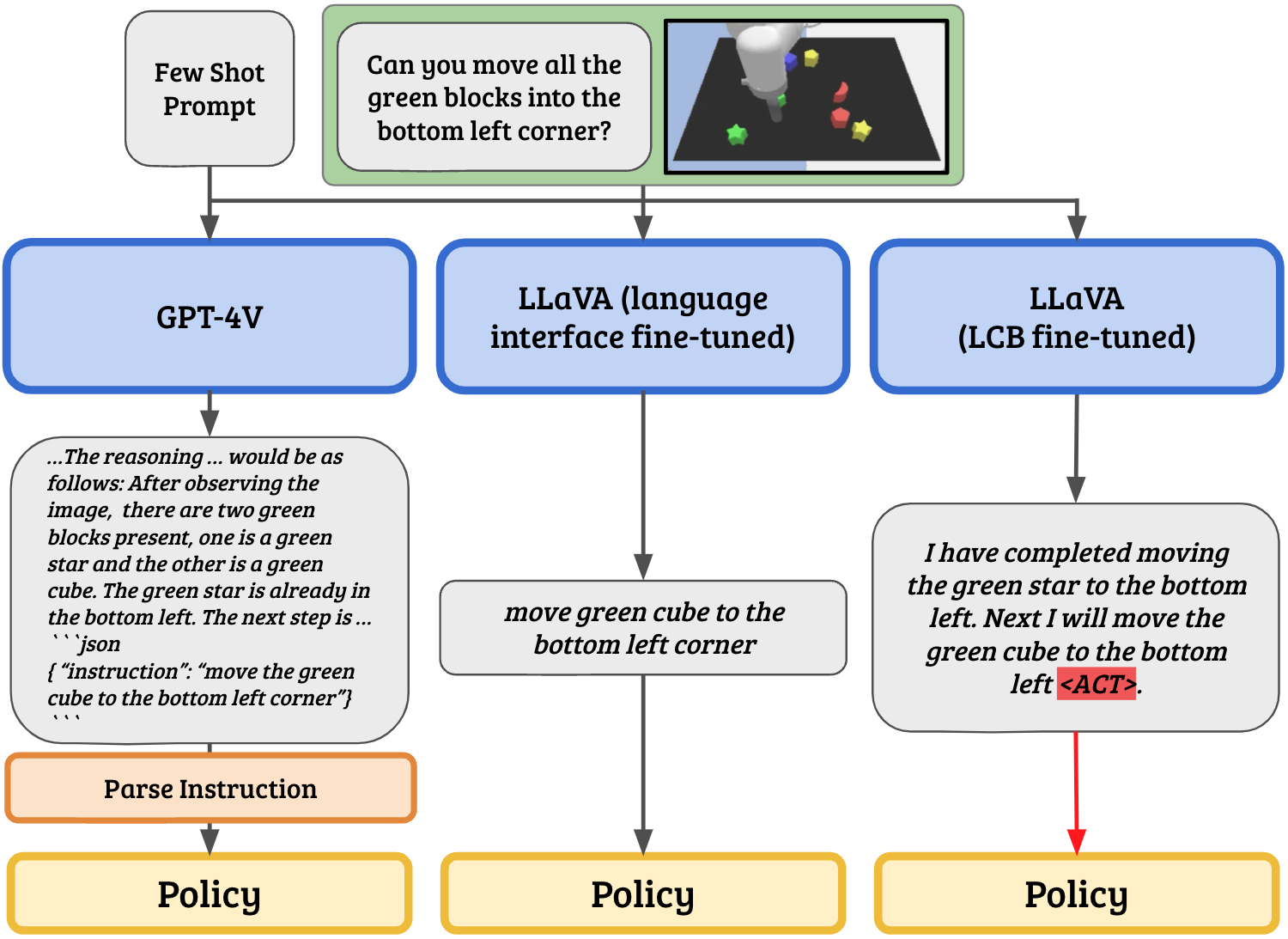}
  \caption{
      A comparison of the flow from a high level language task to the policy for different approaches. \textbf{(Left) LangTable + GPT-4V} requires a prompt to understand the task and desired output format. GPT-4V can provide language reasoning to allow the user to introspect the decision process of the language model, but requires additional parsing to extract the relevant language instruction to provide to the model. \textbf{(Middle) LangTable + LLaVA (Fine-tuned)} fine-tunes the language model to output the exact language instruction as in the training data, effectively acting as a language interface converter. This approach, while effective, removes the chat like capability from the language model. \textbf{(Right) \method} fine-tunes the language model with a chat like interface and action token. The policy is directly conditioned on the latent feature from the action token provided by the model, enabling effective policy conditioning without losing the chat like language model interface.
  }
  \label{fig:lang_compare}
    \vspace*{-3ex}
\end{figure}

%% file: tables/table_calvin_enriched.tex
\begin{table}[h]
  \vspace*{2ex}
\centering
\caption{Task completion rates for various methods on CALVIN\cite{mees2022calvin} long-horizon tasks. All methods were trained exclusively on the ABC split of Calvin with the original language annotations and tested on split D with GPT-4 enriched language annotations, following the RoboFlamingo enriched instruction evaluation setting\cite{liu2024okrobot}. *RF denotes our own training of the RoboFlamingo model on the ABC Calvin split. 3DDA denotes the policy from 3D Diffuser Actor \cite{ke20243d}.}
\label{table:resultcalvin}
\renewcommand{\arraystretch}{1.2} 
\setlength\tabcolsep{8pt} 

\small 
\begin{tabular}{@{}lllll@{}}
\toprule
\multicolumn{1}{c}{\multirow{7}{*}{\rotatebox[origin=c]{90}{\parbox{3cm}{Task Completed\\in a Sequence\\(Success Rate)}}}} & Model & \multicolumn{1}{c}{RF\cite{li2024visionlanguage}} & \multicolumn{1}{c}{3DDA\cite{ke20243d}} & \textbf{\method} \\
\cmidrule(lr){1-5}\\
& \multicolumn{1}{c}{1/5} & 0.620 & 0.652 & \textbf{0.736}\\
& \multicolumn{1}{c}{2/5} & 0.330 & 0.391 & \textbf{0.502}\\
& \multicolumn{1}{c}{3/5} & 0.164 & 0.203 & \textbf{0.285}\\
& \multicolumn{1}{c}{4/5} & 0.086 & 0.117 & \textbf{0.160}\\
& \multicolumn{1}{c}{5/5} & 0.046 & 0.061 & \textbf{0.099}\\
\cmidrule(lr){1-5}
& Avg Len & 0.40 & 1.42 & \textbf{1.78}\\
\bottomrule
\end{tabular}
  \vspace*{-2ex}
\end{table}

%% file: appendix.tex
\onecolumn
\appendices

\section{Dataset Details For Language Table}
\label{append:dataset_details_language_table}

Language table contains a low level text description for each trajectory. We convert this data into chat-like interactions using programmatic templates, common in language based robotics \cite{lynch2021language,shridhar2021cliport,jiang2023vima}. Below in \autoref{fig:data_gen}, we provide the pseudocode with (truncated) examples for how we generate chat question answer pairs for training.
\lstdefinestyle{mystyle}
{
    basicstyle=\small\ttfamily,
    frame=single,
    breaklines,
    columns=fullflexible,
    breakindent=1.2em,
    breakatwhitespace,
    escapeinside={(*}{*)},
}
\begin{figure*}[h!]
\centering
\begin{lstlisting}[style=mystyle,caption={},label=lst:code1,]
QUESTION_LIST = [
    "Can you control the robot to {instruction}?",
    "Can you {instruction}?",
    "Please {instruction}.",
    "Given the current observation, how can you {instruction}?.",
]
def followup():
    start = np.random.choice([None, "first, ", "please, "])
    verb = np.random.choice(["explain", "verbalize"])
    core = np.random.choice([
        "how you would accomplish this task",
        "the desired action",
        "the next step you are going to do",
    ])
    act = np.random.choice(["before acting", "prior to acting"])

    if start is None:
        sentence = verb + " " + core
    else:
        sentence = start + verb + " " + core

    if np.random.rand() > 0.5:
        sentence = sentence + " " + act + "."
    else:
        sentence = act + ", " + sentence + "."

    sentence = sentence[0].upper() + sentence[1:]
    return sentence

def process_instruction(instruction_string, use_extra=True):
    i_string = instruction_string.lower()
    question = np.random.choice(QUESTION_LIST).format(instruction=i_string)
    if use_extra:
        extra_instruction = followup()
        question = question + " " + extra_instruction
    return question

ANSWER_LIST = ["Sure, [ACT].", "[ACT].", "Let's move the robot [ACT]."]
ANSWER_DETAILED_LIST = [
    "I will {detailed_instuction} [ACT].",
    "Sure, I will {detailed_instuction} [ACT].",
    "I should {detailed_instuction} [ACT].",
]
def process_ans_and_ques(instruction_string):
    # sometimes add more details to instruction and privde mode details to the answer
    if np.random.rand() > 0.8:
        question = process_short_horizon_instruction(instruction_string, use_extra=True)
        answer = np.random.choice(ANSWER_LIST)
        answer = answer.format(detailed_instuction=instruction_string)
    else:
        question = process_short_horizon_instruction(instruction_string, use_extra=False)
        answer = np.random.choice(PLANNER_ANSWER_LIST)
    return question, answer

\end{lstlisting}

\caption{Example programmatic generation for \method~training with the original language table data.}
\label{fig:data_gen}
\end{figure*}

\section{GPT-4V Prompting Details}
\label{append:gpt_prompt}

\begin{figure}[h!]
  \centering
  \includegraphics[width=\linewidth]{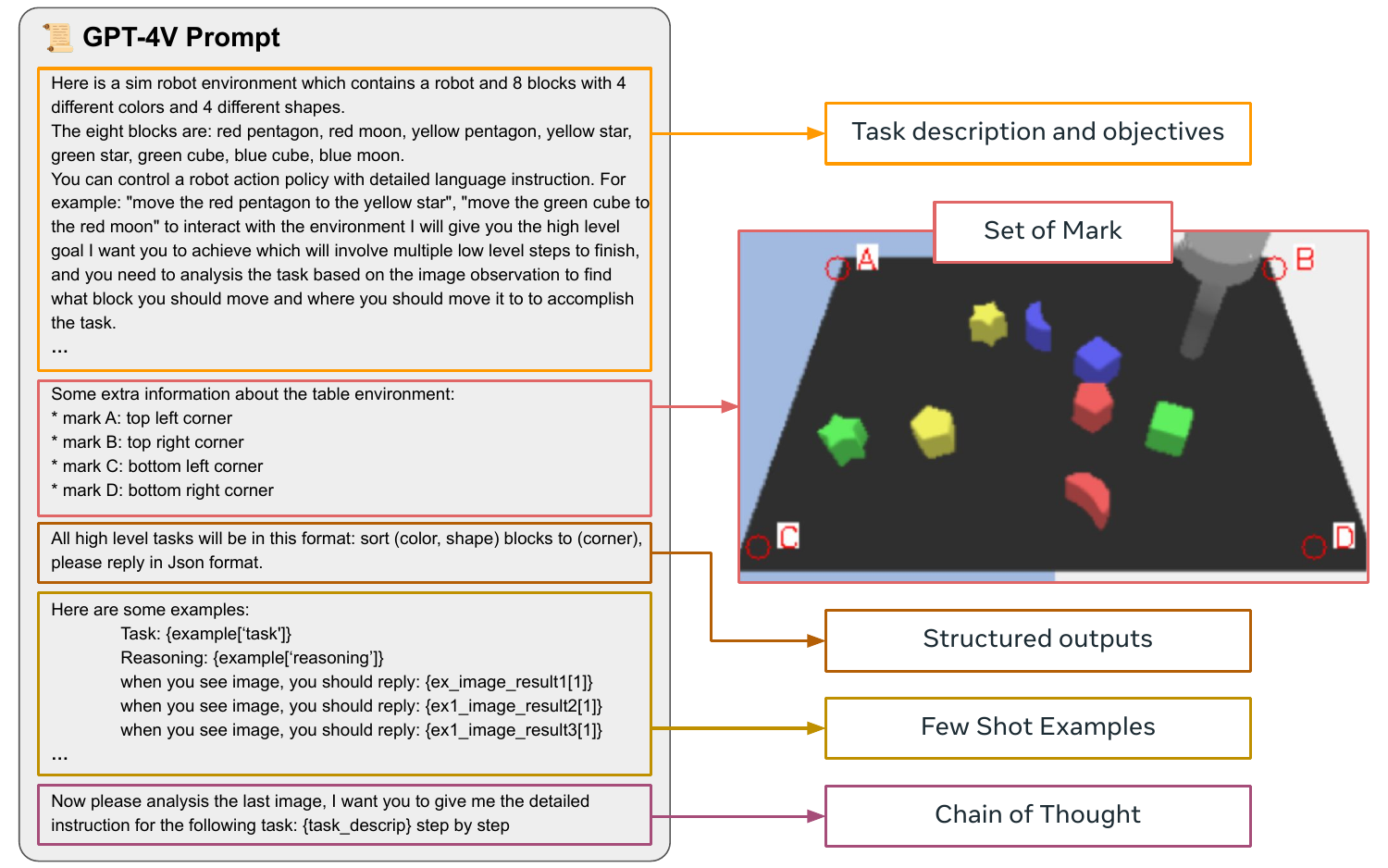}
  \caption{Prompt for using GPT-4V on Language Table as a pretrained VLM. ``..." indicates truncation for brevity, but follows the rest of the text in the section.}
  \label{fig:gpt_prompt}
\end{figure}

GPT-4/GPT-4V is often used zero-shot in robotic applications, due to its strong general understanding. We finetuned our prompt for the language table tasks to achieve the best possible performance, levering prior prompting methods found to improve performance. 
The prompt is seen in detail below in \autoref{fig:gpt_prompt}. We use a comprehensive task prompt as well as Set-of- Mark \cite{yang2023setofmark}, structured outputs, in context examples, and chain of thought prompting \cite{wei2023chainofthought}.

